\documentclass{article}
\usepackage{ijcai26}

\usepackage{times}
\usepackage{soul}
\usepackage{url}
\usepackage[hidelinks]{hyperref}
\usepackage[utf8]{inputenc}
\usepackage[small]{caption}
\usepackage{graphicx}
\usepackage{amsmath}
\usepackage{amsthm}
\usepackage{booktabs}
\usepackage{algorithm}
\usepackage{algorithmic}
\usepackage{amssymb}
\usepackage{amsthm, amsfonts}
\usepackage{amsmath}
\newtheorem{definition}{Definition}

\usepackage{enumitem}
\usepackage[switch]{lineno}

\usepackage{tikz}
\usepackage{forest}
\usepackage{xcolor}

\definecolor{cModel}{HTML}{4A7BA7}      
\definecolor{cAnalytics}{HTML}{6BA368}  
\definecolor{cRepr}{HTML}{C97B63}       
\definecolor{cRoot}{HTML}{555555}       

\title{Beyond Weights and Gradients: A Taxonomy of Federated Learning Messages}

\author{
Alvaro Vargas Guerrero$^{1,2}$
\and
Xinguang Wang$^1$\and
Quang Manh Doan$^1$\and
Guy Nagels$^1$\\
\affiliations
$^1$AIMS lab, Center for Neurosciences, UZ Brussel, Vrije Universiteit Brussel, Brussels, Belgium\\
$^2$Artificial Intelligence Lab, Vrije Universiteit Brussel, Brussels, Belgium\\
\emails
\{Alvaro.Javier.Vargas.Guerrero, Xinguang.Wang, Quang.Manh.Doan, Guy.Nagels\}@vub.be,
}

\begin{document}

\maketitle

\begin{abstract}
Federated Learning is rapidly evolving beyond the exchange of traditional model weights and gradients, yet existing definitions fail to capture the full scope of modern payloads like synthetic data and federated analytics. This paper addresses the gap by proposing a formal mathematical definition of a federated message that accounts for both utility and privacy. We introduce a taxonomy that organizes these exchanges into three categories: model structures, statistical summaries, and data-conditioned representations. By evaluating these groups based on computational demands, communication costs, and privacy risks, we provide a clearer understanding of the trade-offs involved in decentralized training. Our review of 202 recent publications highlights a significant shift since 2021 toward diverse messaging paradigms, signaling a move away from standard deep learning updates toward more specialized information sharing. This framework provides a structured path for future research to optimize federated systems for varying hardware and security requirements.
\end{abstract}

\section{Introduction}

Deep learning is currently an important component in modern artificial intelligence, yet its growth in certain domains is often limited by the need for massive, centralized datasets. Strict privacy laws like the General Data Protection Regulation have made it much harder to collect and share sensitive user data, necessitating methods that respect privacy while enabling robust model training. Federated Learning (FL) addresses this by allowing a central server to coordinate training across multiple clients without ever accessing their raw information. By exchanging only model updates rather than raw data, FL enables privacy-preserving downstream tasks.

The traditional view of FL, pioneered by \cite{mcmahan2016communicationefficientlearningdeepnetworks}, centers on challenges like decentralized data, non-IID distributions, and communication bottlenecks. Early frameworks established a standard where the primary message exchanged between clients and the server consists of deep learning weights or gradients \cite{hard_federated_2019}. Over time, this weight-based paradigm has been scaled to accommodate cross-silo setups, blockchain-governed systems, and various statistical and system heterogeneities \cite{kairouz_advances_2021,liu_recent_2024,alsharif_contemporary_2024,yuan_decentralized_2024}.

However, the field is rapidly evolving beyond the simple exchange of model weights. Modern applications are exploring diverse payloads to improve optimization, communication, and task complexity. Current research utilizes synthetic data as a privacy-preserving proxy \cite{goetz2020federated,Wang2026FAN}, shares intermediate embeddings for vertical federation \cite{yang2023surveyverticalfederatedlearning}, and exchanges graph structures for causal discovery \cite{ng_towards_2022,gao_feddag_2023}. Because the foundational 2016 definition of FL messages no longer captures this breadth, it is necessary to rethink what constitutes a federated payload. In this paper, we propose a formalized definition of a federated message, construct a taxonomy to categorize these modern payloads, and conduct a literature survey to map the current research landscape. By focusing on the message itself, this framework clarifies the trade-offs involved in decentralized training and guides future research toward optimizing federated systems for specific hardware and security requirements.

\section{Related Work}

The structural and algorithmic dimensions of Federated Learning have been extensively mapped by existing literature. Foundational surveys by \cite{yang2019federatedmachinelearningconcept} established the core distinctions between horizontal, vertical, and transfer learning, while subsequent works have categorized the scale of participants (cross-device versus cross-silo) \cite{kairouz_advances_2021} and the topologies of decentralized architectures \cite{yuan_decentralized_2024}. Similarly, recent surveys exploring semi-supervised, self-supervised, and reinforcement learning in federated contexts demonstrate how the system can be adapted for diverse machine learning paradigms \cite{jin_federated_2023,ji_emerging_2024,hatfaludi_foundational_2025}.

Despite these comprehensive overviews, the literature overwhelmingly treats the federated message as a fixed byproduct of the optimization process, invariably assuming the payload is a parameter update. Research into federated communication has primarily focused on efficiency (such as client selection, round reduction, and model compression) rather than questioning the nature of the message itself \cite{zhao2023towards,chen2021communication,ijcai2024p919}. While specialized security literature hints at a broader view by analyzing the exchange of cryptographic keys and noise masks \cite{bonawitz2017practical,behnia2024efficient}, no existing survey has organized these varied types of information into a cohesive taxonomy. Our work fills this gap by shifting the focus away from standard optimization protocols and directly examining the payloads, categorizing them based on their unique computational, semantic, and privacy characteristics.

\section{Definition}

To capture the evolving nature of federated payloads, we step away from overly restrictive statistical boundaries and propose a functional definition. This approach focuses on what a message represents and how it operates within the system, making it applicable to everything from deep learning weights to graph structures and synthetic data.

\subsection{A Functional Definition of a Message}

Consider a federation of $K \ge 2$ parties. Each party $k$ holds a local dataset $D_k$. Let $\theta$ denote an optional shared state provided by the coordinating server or peer network (e.g., a global model, a query prompt, or $\varnothing$ if no prior state exists). 

\begin{definition}[Federated Message]
\label{def:message}
A local mechanism on party $k$ is a function $\mathcal{F}_k$ that processes the local dataset and the shared state. A \textbf{message} $M_k$ is the output of this mechanism transmitted to one or more other parties:
$$M_k = \mathcal{F}_k(D_k, \theta)$$
Importantly, $M_k$ is \emph{data-derived}, meaning its formulation depends non-trivially on $D_k$, but it does not equate to the raw dataset itself ($M_k \neq D_k$).
\end{definition}

\subsection{Utility and Privacy}

Rather than enforcing strict mathematical thresholds that fail to generalize across paradigms, a federated message is evaluated conceptually on two primary axes: its utility to the global task and the privacy it affords the local data.

\begin{definition}[Useful Message]
\label{def:useful}
A message $M_k$ is considered \textbf{useful} if there exists an aggregation rule or a global task where incorporating $M_k$ improves the estimation of a global objective, metric, or model state compared to omitting it. The utility is task-dependent; it may aim to minimize a global loss function, accurately calculate a population statistic, or benchmark global variable relationships.
\end{definition}

\begin{definition}[Private Message]
\label{def:private}
A message $M_k$ is considered \textbf{private} if its transmission structurally or cryptographically bounds the leakage of individual records from $D_k$. This is practically achieved through one or more defense layers: intrinsic masking (e.g., high-dimensional averaging), information-theoretic bounds (e.g., differential privacy noise), or cryptographic wrapping (e.g., secure multiparty computation or homomorphic encryption). 
\end{definition}

\subsection{Multi-Round Composition}

Most federated systems do not rely on a single, isolated exchange. Instead, they operate over $T$ continuous rounds, where the shared state $\theta^{(t)}$ at round $t$ is dynamically updated based on the aggregated messages from prior rounds. This iterative process, known as multi-round composition, fundamentally alters the calculus of both utility and privacy.

From a utility perspective, multi-round composition is often the mechanism that drives the system toward its objective. For gradient updates and low-rank adapters, the continuous sequence of messages is what actually guides the global model toward mathematical convergence. For generative tasks, iterative feedback refines the quality of the synthetic data distribution. In these systems, global utility is an emergent property of the entire transcript of messages over time rather than the result of any single transmission.

Conversely, multi-round composition strictly degrades privacy. Every time a message $M_k^{(t)}$ is transmitted, a fraction of information about the underlying dataset $D_k$ is exposed. Because the shared state $\theta^{(t)}$ adapts based on previous messages, adversaries can analyze the differences between consecutive rounds to isolate and infer individual local data points. This specific vulnerability is frequently exploited in gradient inversion attacks. Therefore, a single-round privacy guarantee is insufficient for continuous federated learning. Robust systems must implement composition accounting (such as tracking a cumulative privacy budget under Differential Privacy) to ensure that the total data leakage over $T$ rounds does not exceed acceptable security thresholds.

\subsection{Applying the Definition to Common Payloads}

To demonstrate the flexibility of this framework, we outline how standard and emerging federated payloads satisfy these definitions in practice.

\begin{itemize}
    \item \textbf{Gradient Update:} In standard FedAvg, the shared state $\theta$ is the global model weights. The mechanism $\mathcal{F}_k$ runs several epochs of backpropagation on $D_k$. The message $M_k = \nabla_\theta \mathcal{L}(D_k, \theta)$ is useful for minimizing global loss, and its privacy is often enhanced via Secure Aggregation to hide individual client contributions.
    \item \textbf{Parameter-Efficient Fine-Tuning (e.g., LoRA):} Here, $\theta$ represents a frozen pre-trained foundational model. The mechanism computes updates only for low-rank adapter matrices $A_k$ and $B_k$. The message $M_k = (A_k, B_k)$ is highly useful for efficiently personalizing large language models, requiring a fraction of the communication bandwidth compared to full gradient updates.
    \item \textbf{Synthetic Dataset:} The mechanism trains a local generative model (such as a GAN or Diffusion model) on $D_k$. The message $M_k$ is a set of generated samples $Z$ that statistically mirror the local distribution. This is useful because it allows the server to compile a proxy dataset for centralized downstream training. Privacy depends heavily on the generalization limits and regularizations of the local generator to prevent memorization.
    \item \textbf{Federated Analytics Query:} The shared state $\theta$ is a specific query (e.g., ``calculate the average age''). The mechanism processes $D_k$ to extract this metric. The message $M_k = \mu_k$ is useful for global monitoring and exploratory data analysis. Privacy is typically enforced by injecting differential privacy noise into the local count or sum before transmission.
    \item \textbf{Embedding (Vertical FL):} In split learning, $\theta$ represents the architecture of the bottom neural network layers. The mechanism passes a raw data sample $x \in D_k$ through these layers. The message $M_k = h(x; \theta)$ is an intermediate activation tensor. It is useful for allowing the server to complete the forward pass without seeing $x$, though it carries higher privacy risks regarding input reconstruction.
    \item \textbf{Graph Structure (e.g., Causal Discovery):} The mechanism calculates conditional independencies or structural scores within $D_k$. The message $M_k$ is an adjacency matrix or a directed acyclic graph (DAG). It is useful for benchmarking methods or creating a global causal structure, representing a vital shift away from optimization-based deep learning payloads toward non-deep parametric models.
\end{itemize}

Having defined how these diverse payloads operate functionally, we now categorize them into three primary groups based on their structural properties and semantic intent.

\section{Taxonomy}

The information exchanged between clients and the server is the most important part of any federated system. To better understand the landscape of modern research, we categorize these messages into three primary groups based on what they represent and how they are used.

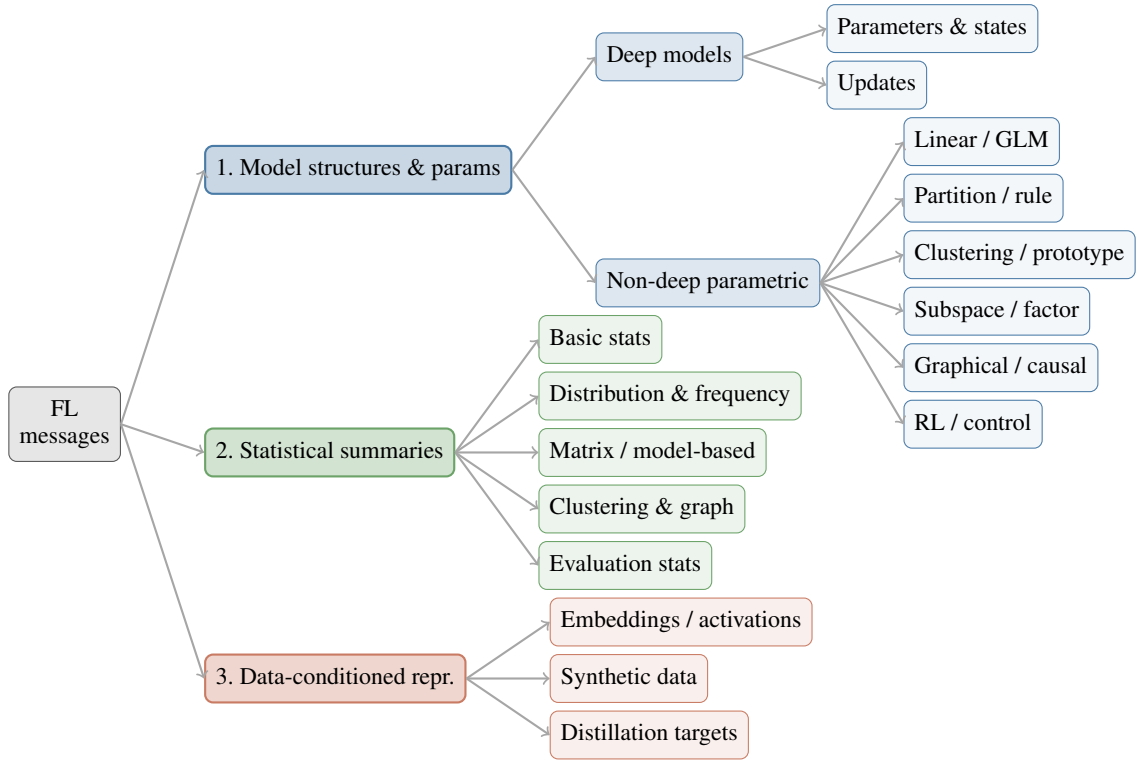
\begin{figure*}[t]
    \centering
\begin{forest}
  for tree={
    grow'=east,
    parent anchor=east,
    child anchor=west,
    draw,
    rounded corners=3pt,
    edge={->, thick, gray!70},
    l sep=11mm,
    s sep=1mm,
    anchor=west,
    align=left,
    font=\small,
    inner sep=4pt,
  },
  [FL\\messages, fill=cRoot!15, draw=cRoot, align=center
    %
    [{1.~Model structures \& params}, fill=cModel!30, draw=cModel, thick
      [Deep models, fill=cModel!18, draw=cModel
        [{Parameters \& states}, fill=cModel!6, draw=cModel]
        [Updates, fill=cModel!6, draw=cModel]
      ]
      [{Non-deep parametric}, fill=cModel!18, draw=cModel
        [{Linear / GLM}, fill=cModel!6, draw=cModel]
        [{Partition / rule}, fill=cModel!6, draw=cModel]
        [{Clustering / prototype}, fill=cModel!6, draw=cModel]
        [{Subspace / factor}, fill=cModel!6, draw=cModel]
        [{Graphical / causal}, fill=cModel!6, draw=cModel]
        [{RL / control}, fill=cModel!6, draw=cModel]
      ]
    ]
    %
    [{2.~Statistical summaries}, fill=cAnalytics!30, draw=cAnalytics, thick
      [Basic stats, fill=cAnalytics!12, draw=cAnalytics]
      [{Distribution \& frequency}, fill=cAnalytics!12, draw=cAnalytics]
      [{Matrix / model-based}, fill=cAnalytics!12, draw=cAnalytics]
      [{Clustering \& graph}, fill=cAnalytics!12, draw=cAnalytics]
      [{Evaluation stats}, fill=cAnalytics!12, draw=cAnalytics]
    ]
    %
    [{3.~Data-conditioned repr.}, fill=cRepr!30, draw=cRepr, thick
      [{Embeddings / activations}, fill=cRepr!12, draw=cRepr]
      [Synthetic data, fill=cRepr!12, draw=cRepr]
      [Distillation targets, fill=cRepr!12, draw=cRepr]
    ]
  ]
\end{forest}
    \caption{Taxonomy of messages exchanged in federated learning.}
    \label{fig:fl-taxonomy}
\end{figure*}

\subsection{Model Structures and Parameters}
\subsubsection{Deep Models}
The most common type of message in federated learning consists of model parameters or their updates. For deep learning, these payloads usually take the form of weight matrices or gradient vectors. In a standard setup like FedAvg \cite{mcmahan2016communicationefficientlearningdeepnetworks}, a client performs several steps of local training and then sends the resulting model state back to the server. These messages are high-dimensional abstractions of the local data. Model weights represent the entire state of a neural network at a specific point in time, while gradients are the directional updates calculated during training that tell the server how to change the global model to reduce error. A key advantage of using such model-based payloads lies in their implicit masking of raw data. Unlike explicit data transmission, these representations do not directly expose original inputs, labels, or sensitive attributes. Instead, they encode information in a distributed and high-dimensional form, which makes direct interpretation difficult.

\subsubsection{Non-deep parametric models}

Non-deep parametric models form a diverse subgroup that focuses on traditional machine learning. Linear and generalized linear models (GLMs) communicate via coefficient vectors, which are very small and offer clear insights into feature relationships \cite{wang_federated_2022,wang_survey_2026}. Tree-based and rule-based models, such as those used in SimFL, involve sharing split statistics or histograms of gradients to build global decision trees without revealing raw feature values \cite{li2020practical}. Clustering models share centroids or prototype sets that summarize data groups, while subspace models like Federated Principal Component Analysis exchange local covariance summaries or eigenvectors to find principal directions of variance across the network \cite{grammenos_federated_2020,elkordy_federated_2023}. For more complex logic, graphical and causal models share structure scores or DAGs to learn the relationships between variables \cite{ng_towards_2022,gao_feddag_2023}. Finally, reinforcement learning models share Q-tables or policy networks to aggregate behaviors learned in different local environments \cite{sutton2018reinforcement}.

\subsection{Statistical Summaries}

Statistical summaries are payloads designed to support federated analytics (FA), focusing on computing global metrics or insights over distributed data without necessarily training a predictive model or maintaining a persistent global parameter vector \cite{wang_federated_2022,wang_survey_2026,elkordy_federated_2023}. In their simplest form, the server issues a query (e.g., “sum of clicks” or “mean of a feature”), and clients respond with local aggregates that are combined into a global answer. Compared to model weights, these messages are usually low dimensional and easy to interpret: sums, counts, means, variances, medians, or percentiles directly describe aspects of the population distribution \cite{wang_federated_2022,wang_survey_2026}. Because of this transparency, statistical summaries are particularly attractive for monitoring, reporting, and exploratory data analysis across silos.

Modern summary payloads extend far beyond basic moments. Distributional representations such as histograms, Cumulative Distribution Function/quantile sketches, and frequency sketches (e.g., Count–Min Sketch for heavy hitters, HyperLogLog for distinct counts) allow clients to send compact payloads while still enabling the server to reconstruct approximate distributions or cardinality estimates. These payloads treat sketching and specialized data structures as core techniques \cite{wang_survey_2026}, supporting a wide range of queries, from simple counts to complex database operations, within a unified framework that is distinct from, yet complementary to, standard federated learning \cite{elkordy_federated_2023}.

\subsection{Data-Conditioned Representations}

Data-conditioned representations consist of objects that are explicit functions of local samples under a given model. A prominent example is the use of embeddings or intermediate activations in split learning and vertical federated learning (VFL). In these settings, each party computes feature vectors for its records using a local encoder; these vectors, often called smashed data, are then sent to a coordinating server or partner party to continue the forward or backward pass \cite{yang2019federatedmachinelearningconcept,yang2023surveyverticalfederatedlearning}.

Synthetic twin data messages take an even more explicit data-like form. Instead of sending gradients or embeddings, clients locally train a generator, such as a Generative Adversarial Network, a Variational Autoencoder, or a gradient-matching procedure. They then upload a small synthetic dataset that approximates the statistical behavior of their real data \cite{goetz2020federated}. Variations of this approach include Federated Knowledge Recycling, which uses locally generated synthetic data for cross-silo environments \cite{lomurno2023fedkr}, FedSyn, which trains a federated GAN to produce global synthetic data \cite{behera2021fedsyn}, and Federated Adversarial Networks (FAN), which leverage generative proxies to robustly handle non-IID distributions and mitigate leakage risks \cite{Wang2026FAN}.

Distillation-based representations complete this group by focusing on model outputs rather than inputs. In knowledge-distillation-based FL, clients evaluate their local models on a shared anchor set, which may be public, synthetic, or jointly maintained. Instead of parameters, they communicate soft labels or logits that convey information about decision boundaries and class relationships \cite{qin2024knowledgedistillationfederatedlearning}.

\subsection{Cross-Group Comparison}

Choosing between these payload types involves a fundamental trade-off between model utility, communication efficiency, and privacy preservation. The choice of a federated payload type is not merely a technical detail but a decision that defines the hardware requirements and security posture of the entire system. By evaluating these message groups across four key dimensions, we can see how the move beyond deep model weights changes the operational landscape of federated learning.

\subsubsection{Local Computation}

The work required from client devices varies significantly depending on the message being produced. Deep model structures demand the highest level of local computation because they require multiple rounds of backpropagation. These intensive energy requirements and long training durations often make deep learning less suitable for constrained devices like low power sensors. In contrast, non-deep parametric models and statistical summaries require very little local work, often involving only simple calculations or a single pass over the local data. Data-conditioned representations present a mixed profile. Generating embeddings or logits only requires a forward pass, which is computationally cheaper than full training. However, creating synthetic data messages can be very demanding, as training a local generator like a GAN requires significant processing power that may exceed the capabilities of many mobile participants.

\subsubsection{Payload Size}

Communication overhead remains a primary bottleneck in federated systems, and the size of the message is the biggest factor. Deep model parameters are typically the largest payloads, with their size scaling directly with the number of network layers and nodes. While compression and sparsification can help, these messages still require substantial bandwidth \cite{konevcny2016federated,li2020federated}. Embedding-based messages can be even larger in high-throughput systems because they scale with both the model dimension and the batch size. Statistical summaries offers the most efficient alternative, as simple counts or means are incredibly small. Synthetic data messages also provide a significant advantage here. Because a few synthetic samples can often represent the local distribution as effectively as a full gradient update, they can achieve order-of-magnitude savings in bandwidth without a large loss in model accuracy \cite{goetz2020federated}.

\subsubsection{Global Aggregation}

Server-side processing time depends on whether the server is performing simple math or active training. For traditional deep models and non-deep parameters, aggregation involves coordinate-wise averaging. While this is mathematically simple, it can become a bottleneck when dealing with billions of parameters or a massive number of clients. Summary messages are the fastest to process, as combining simple statistics is nearly instant. However, data-conditioned representations like synthetic data or distillation targets change the server's role. Instead of just averaging numbers, the server must run a standard training loop on the pooled synthetic data or perform an optimization step to match student and teacher logits. This shift trades higher server computation for lower network traffic. Furthermore, the use of secure protocols like secure aggregation or homomorphic encryption adds a heavy cryptographic load to the server, often turning a simple averaging task into a complex mathematical operation that scales with the number of participants.

\subsubsection{Semantic Granularity}
The depth of knowledge carried by a message determines its utility beyond the primary learning task. Deep model parameters have high semantic granularity because they capture complex patterns across the entire feature space. This richness allows the model to be reused for various downstream tasks or fine-tuning. However, this same density means the message contains extra information that is not strictly necessary for the main goal, which increases the risk of data leakage. Statistical summary messages have the lowest granularity; they only answer a specific question and cannot be used to understand deeper relationships or predict new outcomes. Non-deep models offer a medium level of granularity, providing enough detail for a specific task while staying focused on a narrow set of features. Data-conditioned representations, especially synthetic data, provide high granularity at the sample level. While they may not capture the global decision boundary as directly as model weights, they offer a rich view of the data distribution that is highly useful for data augmentation or semi-supervised tasks.

\subsubsection{Privacy and Defensive Tooling}

The inherent privacy of a federated message depends heavily on its semantic granularity. While early federated systems assumed that deep model weights acted as a natural, high-dimensional shield, subsequent literature has repeatedly demonstrated their vulnerability to gradient inversion and membership inference attacks \cite{zhu2019deep,huang2021evaluating}. Data-conditioned representations, such as intermediate embeddings, carry similar risks because they preserve the precise geometric relationships of the raw input. 

Because gradients and embeddings implicitly carry user-level artifacts, they typically require heavy cryptographic intervention. Secure Multi-party Computation (MPC), specifically Secure Aggregation \cite{bonawitz2017practical}, is commonly deployed for these payloads to ensure the server only observes the combined sum of client updates, thereby hiding individual contributions. Similarly, Homomorphic Encryption (HE) allows the server to aggregate encrypted weight matrices without ever accessing the plaintext \cite{zhang2020batchcrypt}. However, applying HE or MPC to parameter vectors with millions of dimensions introduces immense computational and communication overhead, often bottlenecking the entire system.

In contrast, statistical summary messages are highly transparent. Simple counts or means directly describe the population, meaning repeated or fine-grained queries can easily leak sensitive information about outlier groups \cite{wang_survey_2026,elkordy_federated_2023}. Because these payloads are low-dimensional, they are ideally suited for Differential Privacy (DP). By adding calibrated noise to the aggregate statistics, DP mathematically bounds the risk of individual identification \cite{Abadi2016,geyer2017differentially}. DP is equally critical for synthetic data payloads, where noise is injected during the local generator's training phase to prevent the generative model from inadvertently memorizing and transmitting rare local patterns \cite{goetz2020federated}.

Ultimately, the security of a federated architecture is dictated by the pairing of the payload and the defensive mechanism. The operational friction of protecting deep model weights with HE or MPC is a significant driver behind the field's exploration of alternative messages. By shifting to lightweight analytics protected by DP, or relying on one-shot synthetic data generation, system designers can achieve robust privacy guarantees without the crippling hardware demands of traditional encrypted aggregation. This highlights a critical paradigm shift: the choice of federated payload is no longer just about the learning objective, but about finding a sustainable balance between structural privacy and computational reality.

When evaluating these defensive pairings, it is critical to distinguish between empirical obfuscation and formal privacy guarantees. Data-conditioned representations, such as intermediate embeddings or standard synthetic data generation, typically rely on empirical privacy. They obscure the data heuristically, meaning their security depends entirely on the current limitations of adversarial techniques. If a new, more powerful inversion attack is developed, the empirical shield fails. In contrast, techniques like DP applied to statistical summaries or DP-SGD for model updates provide formal, mathematical bounds \cite{Abadi2016}. These guarantees ensure that the risk of data leakage remains strictly quantified, regardless of an adversary's computational power or auxiliary knowledge. Consequently, while empirical payloads may offer higher immediate semantic utility, systems operating under strict regulatory constraints are increasingly forced to adopt payloads that are natively compatible with formal bounds.

\begin{table}[h]
    \centering
    \caption{Comparison of federated message groups. \\ Abbreviations: \textbf{Local} (Local Computation Load), \textbf{Agg.} (Aggregation Complexity), \textbf{Gran.} (Semantic Granularity).}
    \label{tab:comparison}
    \footnotesize
    \setlength{\tabcolsep}{3pt}
    \begin{tabular}{@{}lccccc@{}}
        \toprule
        \textbf{Group} & \textbf{Size} & \textbf{Local} & \textbf{Agg.} & \textbf{Privacy} & \textbf{Gran.} \\
        \midrule
        Deep models   & High      & High      & Avg.   & Med.  & High    \\
        Non-deep      & Low       & Low       & Avg.   & Low   & Med.    \\
        Summaries   & V.\ low   & V.\ low   & Fast   & Low   & V.\ low \\
        Data-cond.    & Med.      & Med.      & Train. & Low  & High    \\
        \bottomrule
    \end{tabular}
\end{table}

\subsubsection{Synthesis of Trade-offs}
As summarized in Table~\ref{tab:comparison}, no single message type is universally optimal; instead, they represent a complex Pareto frontier of system design. High-granularity payloads, such as deep model updates and data-conditioned representations, maximize utility for complex predictive tasks but demand immense local computation and suffer from severe empirical privacy vulnerabilities. Conversely, statistical summaries offer rapid global aggregation and high compatibility with formal privacy bounds, but sacrifice the semantic depth required for deep learning. Ultimately, the chosen payload dictates the architectural limits of the federation. Mobile, cross-device edge networks are naturally constrained toward lightweight summaries and non-deep models. In contrast, cross-silo institutional networks equipped with robust infrastructure are uniquely positioned to support the heavy cryptographic and computational load required to safely exchange high-dimensional model structures.

\section{Literature Survey \& Results}

To evaluate the shifting landscape of communication payloads, we conducted a targeted literature mapping across a corpus of 202 publications. The collection process relied on a keyword-driven discovery strategy across academic indexing engines and digital repositories, primarily utilizing Google Scholar, arXiv, and IEEE Xplore. The primary search strings were designed to capture alternative messaging paradigms, specifically using queries such as ``Federated Analytics,'' ``Vertical Federated Learning,'' ``VFL,'' and ``Synthetic Data Federated Learning.'' This foundational search was complemented by forward and backward citation snowballing to ensure a comprehensive mapping of cross-disciplinary implementations.

To maintain a strict focus on the nature of the communication payload, papers were subjected to a two-stage screening protocol. The inclusion criterion dictated that a publication must explicitly introduce, modify, or benchmark the structural form of the message transmitted between federated participants. We explicitly excluded papers that focused entirely on hardware routing layers, standard wireless channel optimization, or vanilla convergence proofs that treat the deep parameter update as an unalterable default. During the annotation phase, each of the 202 selected papers was carefully categorized into one of the three pillars of our proposed taxonomy based on the semantic property of its primary message.

\begin{figure}[htbp]
    \centering
    \includegraphics[width=1\linewidth]{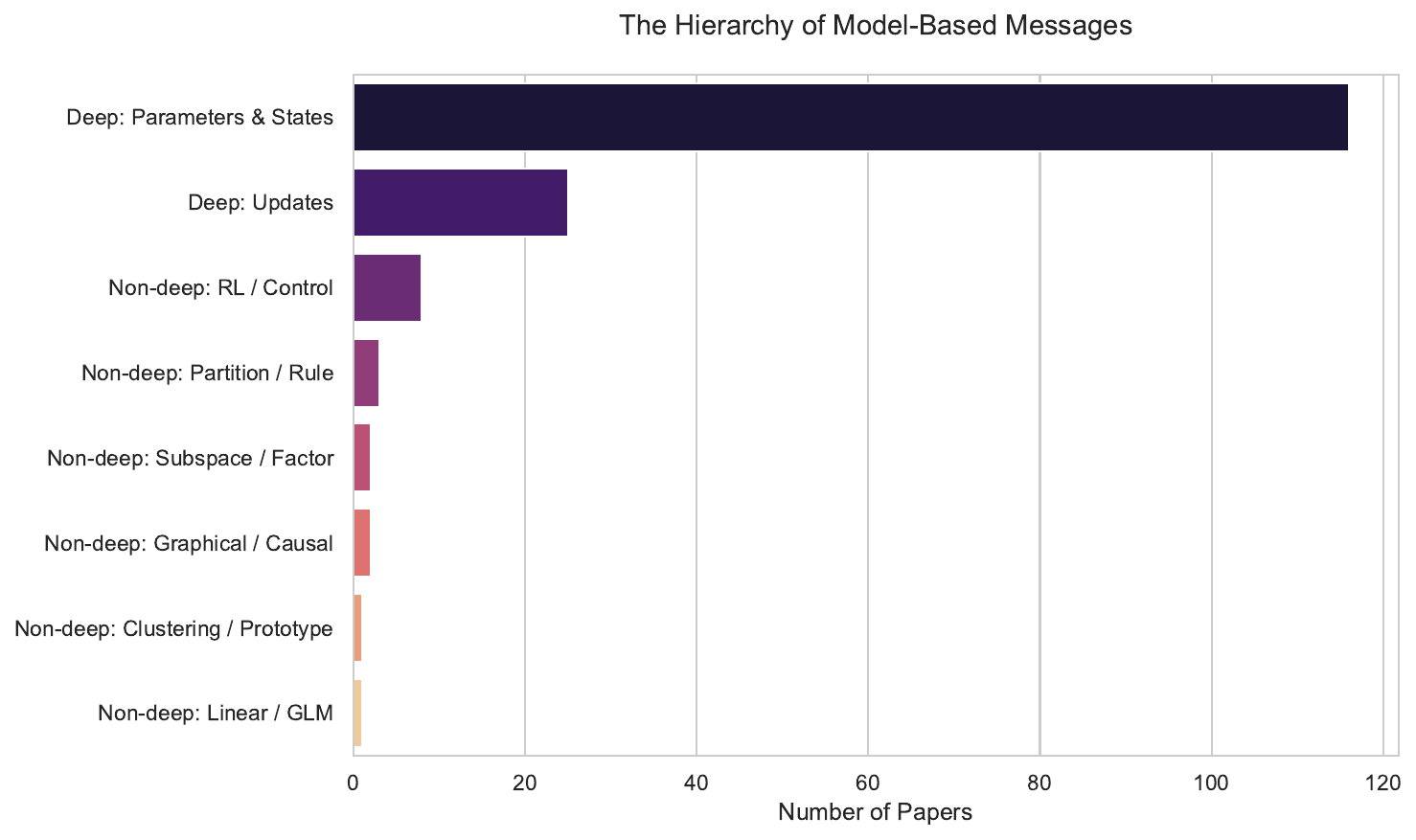}
    \caption{Hierarchy of model-based messages within the "Model Structures \& Params" category.}
    \label{fig:msp_hierarchy}
\end{figure}

\begin{figure*}[htbp]
    \centering
    \includegraphics[width=1\linewidth]{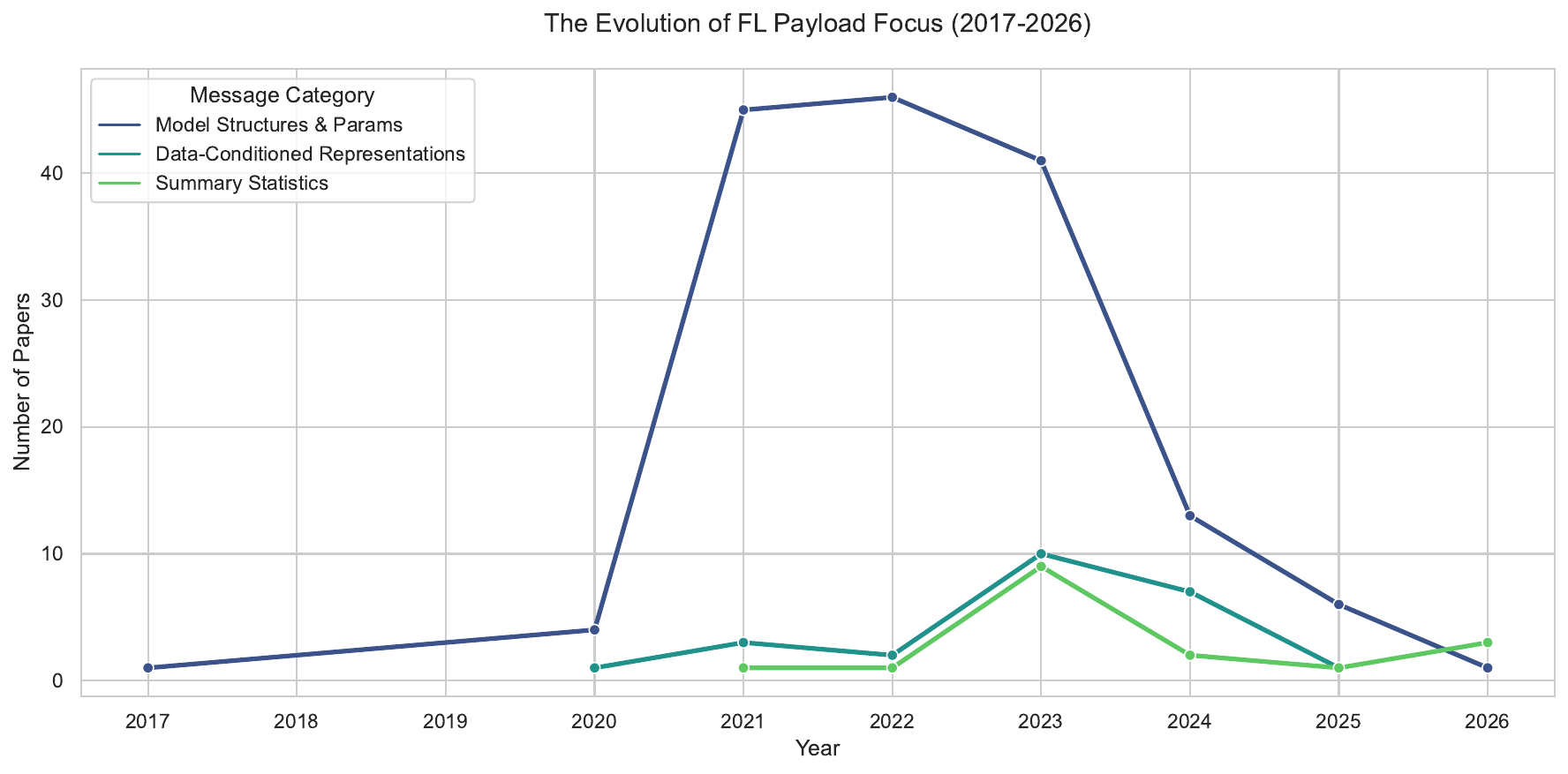}
    \caption{Temporal evolution of Federated Learning research focus (2017--2026).}
    \label{fig:temporal_evolution}
\end{figure*}

\begin{figure}[t]
    \centering
    \includegraphics[width=1\linewidth]{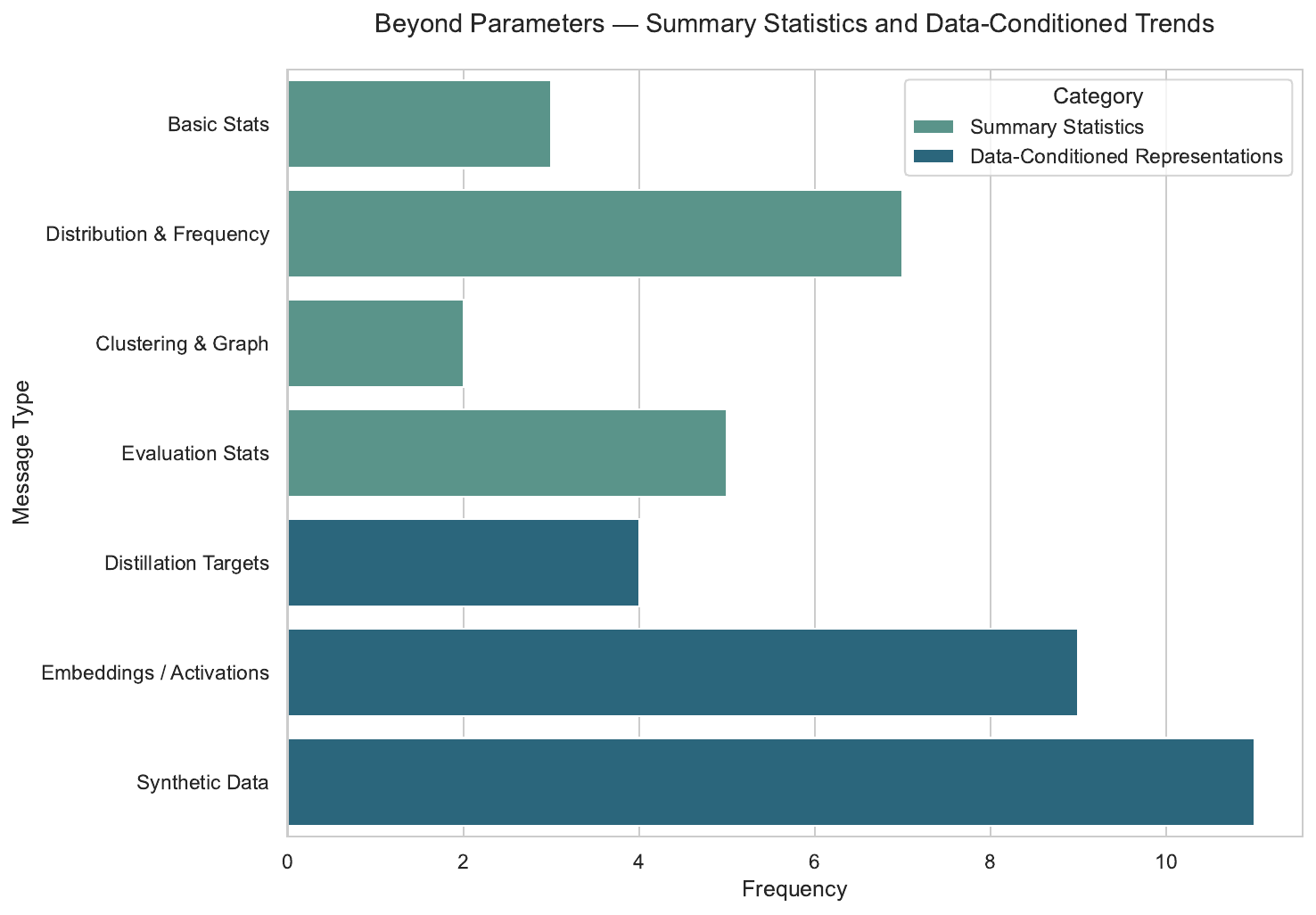}
    \caption{Breakdown of emerging messaging frontiers.}
    \label{fig:frontier_subtypes}
\end{figure}

\begin{figure}[t]
    \centering
    \includegraphics[width=0.8\linewidth]{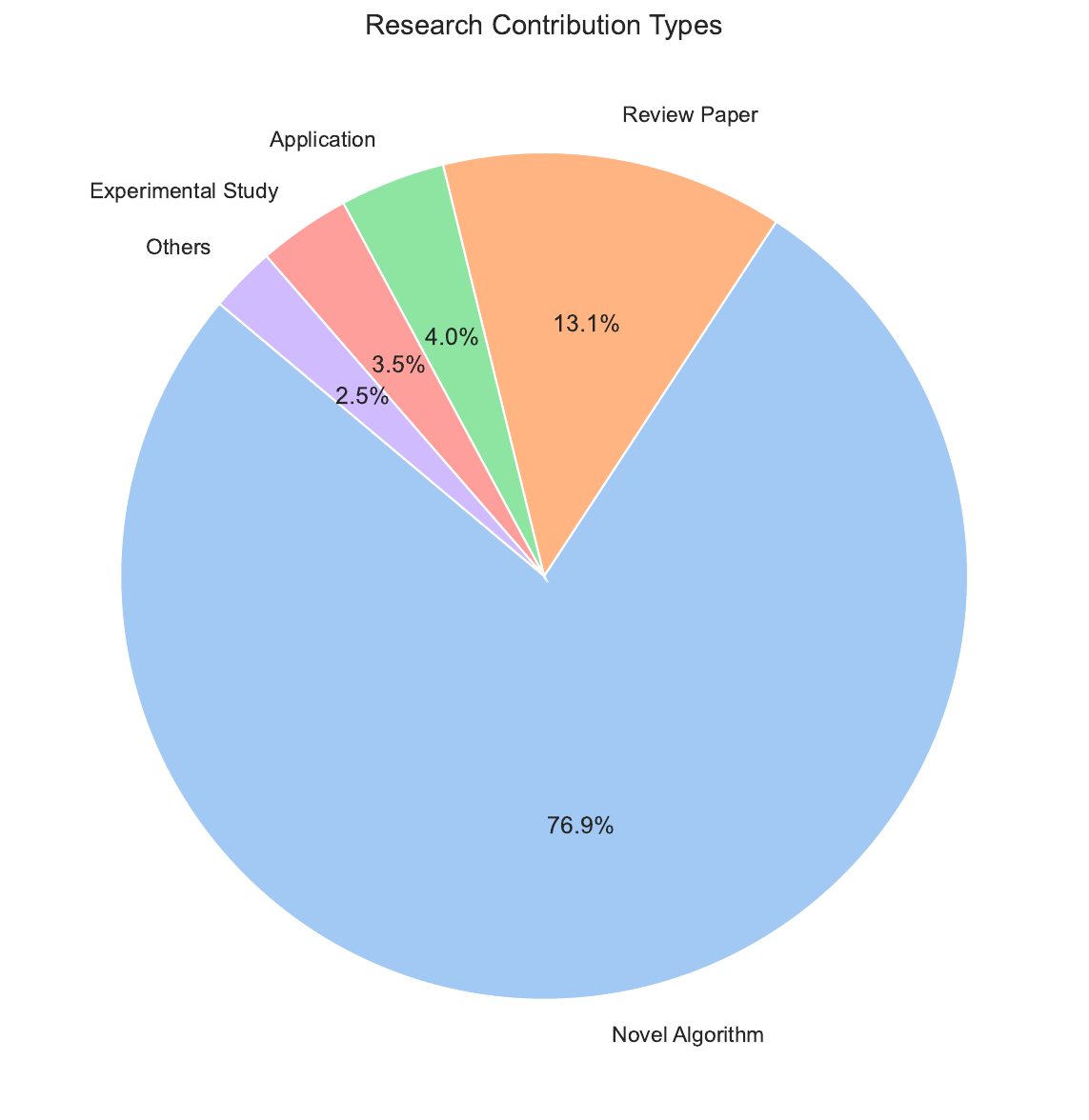}
    \caption{Distribution of primary research contribution types.}
    \label{fig:contribution_types}
\end{figure}
As illustrated in Figure~\ref{fig:msp_hierarchy}, the \textit{Model Structures \& Params} category remains the foundational pillar of the field, dominated by \textit{Deep: Parameters \& States} and \textit{Deep: Updates}. 
However, this group has increasingly diversified to include non-deep parametric models, such as those for \textit{RL/Control} and \textit{Subspace/Factor} analysis, reflecting a broader application scope beyond standard neural networks. 
Figure~\ref{fig:temporal_evolution} captures the temporal evolution of these research priorities, revealing a critical diversification of the FL ecosystem beginning around 2021. 
While traditional model-based approaches persist, there is a marked upward trend in \textit{Data-Conditioned Representations} and \textit{Statistical Summaries}, as seen in Figure~\ref{fig:frontier_subtypes}. 
This shift signals an advancement into more sophisticated paradigms: \textit{Data-Conditioned Representations} now fuel progress in VFL via \textit{Embeddings/Activations}, while \textit{Synthetic Data} and \textit{Distillation Targets} provide robust alternatives to raw weight transmission. 
Concurrently, \textit{Summary Statistics} has experienced a resurgence, driven by its capacity for one-shot FL and the use of lightweight payloads \cite{turazza2026gaussianheadoflfamilyoneshot} such as \textit{Distribution \& Frequency} statistics and \textit{Basic Stats}, to minimize communication overhead. 
Finally, Figure~\ref{fig:contribution_types} highlights that the field is fundamentally driven by methodological innovation, with \textit{Novel Algorithms} constituting approximately 76.9\% of the research, far outpacing specific applications or experimental studies.

\section {Future Work}
As this research is currently a work in progress, our immediate priority is to expand the literature search across a wider range of academic databases. While the current survey of 202 papers provides a strong signal of shifting trends, a more exhaustive query of sources like Scopus and Web of Science will ensure the taxonomy's empirical foundation is as robust as possible. We plan to utilize these findings to create a more granular map of which message types are gaining the most traction in specific industries, such as the preference for analytics in finance versus embeddings in healthcare.

Beyond the literature review, we intend to develop a standardized benchmarking framework to empirically evaluate the trade-offs described in our taxonomy. Currently, choosing a payload type is often a matter of intuition or following convention. We will design an experimental suite that isolates specific tasks, such as predicting tabular health outcomes versus generating synthetic imaging proxies, to test model weights, statistical summaries, and distillation targets under uniform hardware limits. By measuring the exact computational overhead of encrypting a gradient update versus injecting DP noise into an analytical query, we aim to provide system designers with a strict, quantitative decision matrix for matching payloads to specific security postures.

\section{Conclusion}
Federated Learning is maturing beyond its original focus on simple weight-based optimization. As the field expands to include complex tasks like causal discovery and reinforcement learning, the traditional definition of a federated message has become a bottleneck for both research and practice. In this paper, we have proposed a more inclusive, formal definition of the federated message that accounts for the wide variety of modern payloads. By categorizing these messages into model structures, statistical summaries, and data-conditioned representations, we have created a structured map for the diverse information being shared across decentralized networks today.

Our literature survey confirms that the research landscape is diversifying rapidly. While deep learning parameters remain the dominant form of communication, the rise of data-conditioned representations and analytics-driven approaches highlights a growing need for communication efficiency and task-specific granularity. The choice of a message type is no longer just a technical detail: it is a strategic decision that affects the privacy, speed, and hardware requirements of the entire federation. We hope that this taxonomy and definition provide a clearer path for researchers to move beyond the constraints of traditional updates and explore the full potential of federated communication. \newpage

\section*{Acknowledgments}
This work was supported by the Research Foundation Flanders (FWO) under Senior Research Project No. G0A2Q25N.
Guy Nagels is a Senior Clinical Research Fellow of the Fonds Wetenschappelijk Onderzoek Vlaanderen (FWO.be, grant 1805625N)

\bibliographystyle{named}
\bibliography{ijcai26}

\end{document}